\documentclass{article}
\usepackage{kr}

\usepackage{times}
\usepackage{soul}
\usepackage{url}
\usepackage[hidelinks]{hyperref}
\usepackage[utf8]{inputenc}
\usepackage[small]{caption}
\usepackage{graphicx}
\usepackage{amsmath}
\usepackage{amsthm}
\usepackage{booktabs}
\usepackage{algorithm}
\usepackage{algorithmic}

\usepackage{pgfplots}

\usepackage{color}
\usepackage{amsfonts} 
\usepackage{subcaption}

\title{TaxoKnow: Taxonomy as Prior Knowledge in the Loss Function of Multi-class Classification}

\author{%
    Author name
    \affiliations
    Affiliation
    \emails
    email@example.com    % email
}

\author{%
Mohsen Pourvali\and %$^1$\and
Yao Meng \and %$^1$\and
Chen Sheng \and %$^1$\and
Yangzhou Du \\ %$^1$ \\
\affiliations
Lenovo AI Lab, Beijing, China\\
\emails
\{mpourvali, mengyao1, shengchen1, duyz1\}@lenovo.com,
}
\begin{document}

\maketitle

\begin{abstract}
  In this paper, we investigate the effectiveness of integrating a hierarchical taxonomy of labels as prior knowledge into the learning algorithm of a flat classifier. We introduce two methods to integrate the hierarchical taxonomy as an explicit regularizer into the loss function of learning algorithms. By reasoning on a hierarchical taxonomy, a neural network alleviates its output distributions over the classes, allowing conditioning on upper concepts for a minority class. We limit ourselves to the flat classification task and provide our experimental results on two industrial in-house datasets and two public benchmarks, RCV1 and Amazon product reviews. Our obtained results show the significant effect of a taxonomy in increasing the performance of a learner in semi-supervised multi-class classification and the considerable results obtained in a fully supervised fashion.
\end{abstract}

\section{Introduction}

Large Language Models (LLM), such as GPT-3 \cite{floridi2020gpt}, have made significant advances in Natural Language Processing (NLP). In general, pre-training, where a model first trains on massive amounts of data before being fine-tuned for a specific task, has proven to be an efficient technique for improving the performance of a wide range of language tasks \cite{min2021recent}.  For example, BERT \cite{devlin2018bert} is a pre-trained transformer-based encoder model that can be fine-tuned for various NLP tasks, such as sentence classification, question answering, and named entity recognition. In fact, large language models have shown a so-called few-shot learning capability to be efficiently adapted to downstream tasks.

\begin{figure}
	\centering
	\def\svgwidth{270pt}
	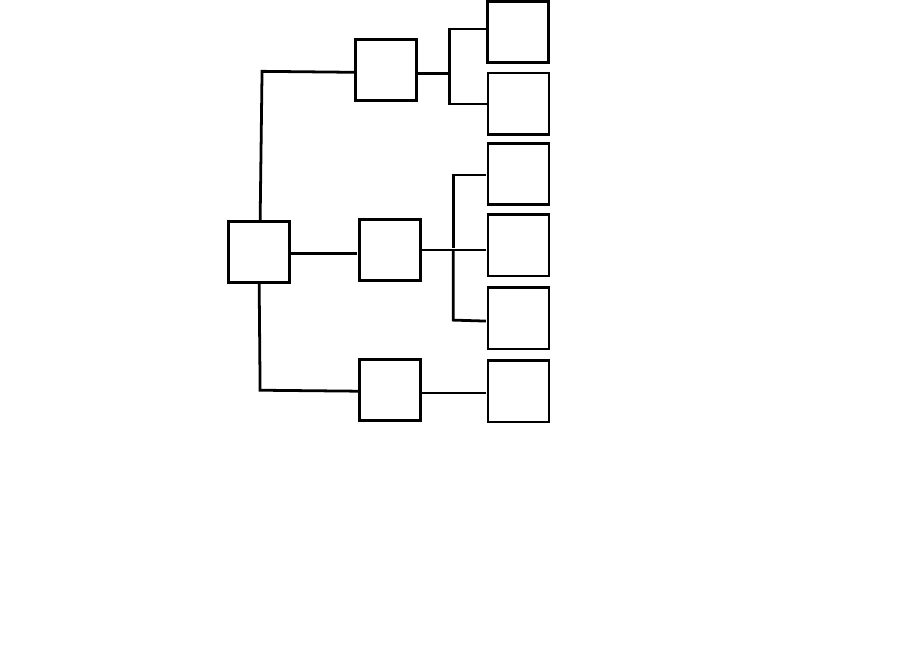
	\caption{A symbolic representation/sentence for node $a_{2}$ in a higher level $l_{2}$ of a hierarchical taxonomy for multi-class classification}
	\label{fig:taxonomy}
\end{figure}

If we break down the architecture of LLMs, we can categorize their components into two general concepts: Deep Neural Network (DNN) as a part of Machine Learning (ML), and Trained Data. Despite all the advantages of LLMs, they come with some limitations. Starting from the very beginning, machine learning has its own limitations, from supervised ML which heavily relies on large amounts of human-labeled data to Reinforcement Learning (RL) which requires a very large number of interactions between the agent and the environment. The brittleness of deep learning systems is largely due to machine learning models being based on the independent and identically distributed (i.i.d.) assumption, which is not a realistic assumption in the real world. Moreover, compared to human capabilities, DNNs still lack in various aspects, such as Adaptability, Generalizability, Robustness, Explainability, Abstraction, Common sense, and Causal reasoning. In general, Multi-Layer Perceptrons (MLPs) are good at generalizing within the space of training examples, but they perform poorly at generalizing outside the space of training examples, and this limitation is not improved even by adding more layers. So, the question is, what can be done? Can increasing the size of trained data solve these shortcomings?

Another shortcoming, which is not addressed by simply using more data, is \textit{curve fitting} \cite{pearl2019limitations}, mapping inputs to outputs. If our systems rely solely on curve-fitting and statistical approximation, their inferences will necessarily be shallow. Instead of inducing a more abstract and causal understanding of the world, they try to approximate the statistical curves of how words are used to infer how the world works. Consider the following example from GPT-2:

\begin{align*}
\textit{``\textbf{A good way to light a fire is} to use a pint-sized furnace.''}
\end{align*}
%\textit{``\textbf{A good way to light a fire is} to use a pint-sized furnace.''}

It clearly shows a lack of understanding of the nature of fire by the model, as it is essentially based on the statistical occurrence of words next to each other, without an inherent cognitive model. 

Let us take a step back and explore another approach to training a machine, which is \textit{Symbolic Machine} learning. A symbolic machine combines a sophisticated reasoner with a large-scale knowledge base. One well-known example of a symbolic machine is CYC \footnote{\url{www.cyc.com}}, which represents the single largest effort in the history of AI to create common-sense knowledge in a machine-interpretable form. CYC was launched in 1984 by Doug Lenat and required thousands of person-years of effort to capture facts about psychology, politics, economics, biology, and various other domains in a precise logical form. One famous test of CYC is the Romeo and Juliet quiz, in which CYC demonstrates an internal distillation of a complex scenario and provides an example of rich cognition. However, despite the extensive efforts put into CYC, it falls short compared to the remarkable results achieved by transformers and GPT-2, even without explicit knowledge engineering.

What Gary Marcus \cite{marcus2020next} believes is that symbol manipulation could be the solution, particularly for extrapolating beyond a training regime. Symbol manipulation, specifically the machinery of operations over variables, offers a natural albeit incomplete solution to the challenge of extrapolating beyond a training regime. It also provides a clear basis for representing structured representations (such as the tree structures foundational to generative linguistics) and records of individuals and their properties. It, along with a satisfactory framework for learning, can bring a hybrid approach that combines the best of both worlds: the ability to learn from large-scale datasets and the capacity to represent abstract concepts. The power of combining statistical and symbolic artificial intelligence techniques to accelerate learning and improve transparency is exemplified by the work of MIT-IBM Watson AI Lab and DeepMind \cite{mao2019neuro}.

In this work, we aim to integrate abstract/prior knowledge (Hierarchical Taxonomy of labels) into the structure of machine learning. As one of our contributions, we leverage symbolic manipulation to represent the taxonomy. According to Henry Kautz's proposal on Neural-Symbolic Computing (NSC) \footnote{\url{https://www.cs.rochester.edu/u/kautz/talks/index.html}}\cite{garcez2023neurosymbolic}, our work can be categorized as type 5; $\small{NOURO_{SYMBOLIC}}$; a tightly-coupled neural-symbolic system where a symbolic logic rule is mapped onto a distributed representation (an embedding) and acts as a soft-constraint (a regularizer) on the network’s loss function.
Additionally, we combine type 5 with a method from type 1; \textit{\MakeUppercase{symbolic neuro symbolic}}; which involves standard deep learning in which input and output of a neural network can be made of symbols. Our target is an imbalanced classification problem where we have a Hierarchical Taxonomy of labels as our prior knowledge.

Many real-world classification problems exhibit imbalanced class distributions, such as having 25 fraudulent transactions among 1,000,000 normal transactions in a financial security dataset of a reputable bank \cite{hasanin2019severely}.
In current fully supervised classification tasks, models are trained on labeled datasets where labels are primarily injected into the objective function (e.g., cross-entropy) as prior knowledge. These labels typically originate from a larger hierarchical taxonomy, allowing for comprehensive reasoning over the labels. 
%For example, in a classification task with 6 classes \textit{\{screen\_broken, Hardware\_upgrade, Bios\_update, Windows\_installation, T-shirt, Sweater\}} a higher level of concept would be \textit{\{(screen\_broken) $->$ Mobile\_service, (Hardware\_upgrade, Bios\_update, Windows\_installation)$->$ Computer\_service, (T-shirt, Sweater)$->$ Clothing\}}. 

Labels in Machine Learning (ML), especially in supervised ML, play an important role. Among the three main components of supervised ML, data and models have received the most attention in active research \cite{bagherinezhad2018label}. However, labels often present challenges. Despite the human cost required for labeling, labels are frequently incomplete, ambiguous, and redundant. Using a hierarchical taxonomy for labels can provide more information that leads to improved labels and ultimately enhances model quality in supervised learning, and even yields further gains in semi-supervised learning. 
%Existing research has made efforts to integrate such the additional information directly/indirectly into ML in terms of; data, model, loss function, etc. 

In this paper, our aim is to inject a hierarchical taxonomy of classes into the loss function of the learning algorithm for text classification, which can be scaled to other data domains. We introduce two methods to represent and incorporate the hierarchical taxonomy. The first method (Section \ref{sec:symbol}) represents the taxonomy as constraints in Boolean logic. For example, Figure \ref{fig:taxonomy} illustrates a hierarchical taxonomy for class labels, where leaves at level $l_{1}$ indicate the actual class labels used in the loss function (e.g., cross-entropy), and nodes at a higher level $l_{2}$ indicate a higher level of conceptualization for the labels, which are typically not used in the classification algorithm. Our goal is to use the higher levels of the taxonomy to constrain the distribution of neural network outputs. Similar to \cite{xu2018semantic}, we enhance neural networks with the ability to learn how to make predictions while adhering to these constraints, leveraging symbolic knowledge to improve learning performance. 
The second method (Section \ref{sec:gcn}) involves using Graph Convolutional Networks (GNN) to represent and incorporate the hierarchical taxonomy into the loss function. Our experimental results for both methods demonstrate the significant effect of higher levels of the hierarchical taxonomy in alleviating the unequal distribution of classes in severely imbalanced classification problems. 

Our contributions in this paper focus on flat/general classification, referring to the standard multi-class classification problem. This differs from hierarchical classification, where the class set to be predicted is organized into a class hierarchy, typically represented as a tree or a Directed Acyclic Graph (DAG).

\section{Related Work}
\label{sec:relatedwork}

\subsubsection{Imbalanced Classification:} Approaches for dealing with imbalanced classification problems can be categorized into three groups: data-level approaches, algorithm-level techniques, and hybrid methods \cite{johnson2019survey}. 
Data-level approaches aim to address the unequal distribution of classes by employing sampling techniques such as over-sampling the minority class or under-sampling the majority class. However, under-sampling may result in the loss of important information for the model to learn from, while over-sampling can increase training time and lead to overfitting \cite{johnson2019survey}.
Algorithm-level techniques, on the other hand, adjust the learning or decision process to give more importance to the minority class. Hybrid methods combine data-level and algorithm-level approaches in various ways to tackle the class imbalance problem \cite{seiffert2009rusboost,chen2021hybrid}.

\subsubsection{Taxonomy-aware Classification:} The use of hierarchical concepts in classification has been explored in various studies. For example, \cite{brust2020integrating} leverages a publicly available hierarchy like WordNet to integrate additional domain knowledge into classification. Exploiting taxonomy to enhance model quality is not a new concept, as \cite{cai2007exploiting} directly incorporates taxonomic information into the model architecture.

Existing works on integrating taxonomy into machine learning can generally be grouped into two approaches. The first approach involves \textit{indirectly} incorporating taxonomy information into the ML model, such as label expansion \cite{li2017learning}. The second approach focuses on \textit{directly} integrating taxonomy information into the model architecture \cite{karamanolakis2020txtract,jenkins2021taxonomy,ong2022image}. 
%The hierarchical nature of objects in real world provides an initial categorization for them. 
\cite{ong2022image} demonstrates that using taxonomy information of plant species can alleviate class sparsity issues when optimizing for a large number of classes.
In the domain of semi-supervised learning, \cite{su2021semi} proposes techniques for incorporating coarse taxonomic labels to train image classifiers in fine-grained domains.

While previous works have explored the effectiveness of label taxonomies in hierarchical classification, our paper emphasizes the positive impact of hierarchical taxonomies in flat classification problems. To the best of our knowledge, our work is the first to propose injecting hierarchical taxonomies of labels as prior knowledge into flat classification problems. Our approach falls under algorithm-level techniques for addressing imbalanced classification problems, as we directly inject a hierarchical taxonomy of class labels as prior knowledge into the existing loss function (i.e., data-driven) of a deep neural network.

% and its main effort is to show that the hierarchical taxonomy of classes for an imbalanced dataset can lead classification learner to account for class imbalance. We highlight the positive effect of hierarchical taxonomies in the problem of imbalance flat classification which can be used as explicit regularization terms along with any other approach to deal with imbalanced data.

\section{Proposed Methods}
We propose two approaches to represent and integrate the hierarchical taxonomy as prior knowledge into the loss function of a learning algorithm.

\subsection{Symbolic-based Approach}
\label{sec:symbol}
To integrate the hierarchical taxonomy of the classes into the loss function, we first represent the taxonomy as symbolic logical constraints. Building on the work of \cite{xu2018semantic} we derive a differentiable semantic loss function that captures how well the neural network satisfies the constraints on its output.

\subsubsection{General Notation.} We employ concepts in propositional logic to formally define taxonomy and semantic loss. Boolean variables are written in uppercase letters ($X,Y$), and their instantiation ($X=0$ or $X=1$) are written in lowercase ($x,y$). We write sets of variables in bold uppercase ($\textbf{X},\textbf{Y}$), and their joint instantiation in bold lowercase ($\textbf{x}, \textbf{y}$). A literal is a variable ($X$) or its negation ($\neg X$). A logical sentence ($\alpha$ or $\beta$) is created by variables and logical connectives ($\wedge$, $\vee$, etc.), and is also called a formula or constraint. A state $\textbf{x}$ satisfies a sentence $\alpha$, denoted as $\textbf{x} \models \alpha$, if the sentence evaluates
to be true in that world, as defined in the usual way.
The output vector of a neural network is denoted by $\textsf{p}$, where each value in $\textsf{p}$ represents a probability of an output in $[0,1]$.
The output vector of a set of sentences is denoted by $\textsf{s}$, where each value in $\textsf{s}$ represents a satisfaction value in $[0,1]$.

\subsubsection{Taxonomy.}
Each level of concepts in a taxonomy is denoted as $l_{i}, i \in [1,K]$, where $K$ is node-based length of the taxonomy and $l_{1}$ indicates the leaves of the taxonomy, which is associated with the class labels. Each node in taxonomy except nodes in the leaves is denoted as $a_{i}$. For instance, in Figure \ref{fig:taxonomy}, for a taxonomy used in multi-class classification, sentence $\alpha$ states that for a set of indicators $\textbf{X}=\{X_{1},..,X_{6}\}$, one and exactly one of $X_{3}, X_{4}, X_{5}$ must be true, while the rest must be false. This statement indeed represents node $a_{2}$ of the taxonomy in terms of its children/variables ($X_{3}$, $X_{4}$, $X_{5}$). 
To represent hierarchical nature of the taxonomy, a set of variables $\textbf{B}=\{B_{1},B_{2},..,B_{K-1}\}$ is defined over the taxonomy levels. 
$B_1, ..., B_{K-1}$ correspond to the variables of each non-leaf node in the taxonomy tree, where each variable of $\textbf{B}$ corresponds to a set of one-hot vectors $b_j$, e.g., $b_1$ and $b_2$ correspond to level 1 and level 2, as shown in Figure \ref{fig:illus_super} and Figure \ref{fig:illus_semi}. $b_j$ corresponds to one-hot vector over ${a_0, a_1, ..., a_m}$, where $m$ is number of nodes in level $l_j$, e.g., as it is shown in Figure \ref{fig:taxonomy} ${a_1, a_2, a_3}$ for level $l_2$.
A logical sentence $\beta$ is created from variables $\textbf{B}$ and logical connective $\wedge$. 
For a given taxonomy there would be a sentence $\delta_{ij}$ corresponding to the node $a_i$ (i.e., propositional logic) which all sentences for each level $l_j$ are stored in $d_j$ ($d_1$ and $d_2$ as it's shown in Figure 2), and sentence $\alpha$ is defined over ${d_1, d_2, ..., d_{K-1}}$.

\subsubsection{Semantic Loss.}
The semantic loss $L^s(\alpha,\beta,\textsf{p},\textsf{s})$ is defined as a function of sentences ($\alpha$,$\beta$) in propositional logic, which is defined over variables $\textbf{X}=\{X_{1}, X_{2},..,X_{n}\}$ and $\textbf{B}=\{B_{1},B_{2},..,B_{K-1}\}$, a vector of probabilities $\textsf{p}$ for variables $\textbf{X}$, and a satisfaction vector $\textsf{s}$ for variables $\textbf{B}=\{B_{1},B_{2},..,B_{K-1}\}$. The element $\textsf{p}_{i}$ denotes the predicted probability of variable $X_{i}$, corresponding to a single output of the neural network. The element $\textsf{s}_{i}$ represents the satisfaction score of variable $B_{i}$, corresponding to the output of a sentence $\alpha$. 
Similar to \cite{xu2018semantic}, we provide two examples of integrating semantic loss $L^s$ into an existing loss function as an additional regularization term, in both supervised and semi-supervised manners. Specifically, with a weight $w$, the new loss becomes

\begin{align*}
existing\_loss + w \cdot semantic\_loss.
\end{align*}

\subsubsection{Supervised-based Definition.}
In the Supervised-based definition, we assume that all the training dataset is labeled, and the hierarchical taxonomy is complete, meaning that for labeled class, all the upper parents are known. Formally, for a class label $cl_{i}$, its $K-1$ upper concepts in the taxonomy are given. With this assumption, let $\textsf{p}$ be a vector of probabilities, one for each variable in $\textbf{X}$, let $\alpha$ be a sentence over $\textbf{X}$, and $\beta$ be a sentence over $\textbf{B}$.

\begin{small}
	\begin{equation}
	\label{eq:super}
	L^s(\alpha,\beta,\textsf{p},\textsf{s}) \propto - \log \prod_{\textbf{y} \models \beta} \sum_{\textbf{x} \models \alpha} \prod_{i:\textbf{x} \models X_{i}} \textsf{p}_{i} \prod_{i:\textbf{x} \models \neg X_{i}} (1- \textsf{p}_{i})
	\end{equation}
\end{small}%

\begin{figure}[t]
	\centering
	\def\svgwidth{270pt}
	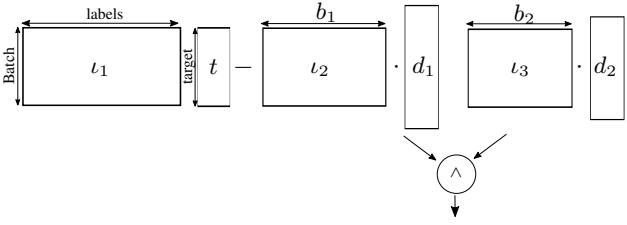
	\caption{An illustration for supervised semantic loss. $\iota_1$ is the matrix of model output for a batch, $\iota_2$ is the matrix of one-hot vectors over nodes in level $l_2$ in the batch, and $\iota_3$ is the matrix of one-hot vectors over nodes in level $l_3$ in the batch. $|b_i|=|d_i|$ since each element in one-hot vector $b$ is corresponding to a sentence $\delta$ in $d$.}
	\label{fig:illus_super}
\end{figure}

Equation 1 1 represents the hierarchical taxonomy as a logical constraint. By applying the negative logarithm, we enforce the training model to satisfy the constraint. 
Figure \ref{fig:illus_super} provides an illustration of Equation \ref{eq:super}, including target labels, training batch, and SDDs. The target labels $t$ are utilized in an existing data-driven loss function, such as cross entropy.

\subsubsection{Semi-supervised-based Definition.}
There is a growing interest in utilizing unlabeled data to augment the predictive power of classifiers. In this section, we demonstrate the integration of a hierarchical taxonomy with unlabeled data.
In the Semi-supervised-based definition, the assumption is that there is unlabeled data and the hierarchical taxonomy is not complete. The semantic loss is defined for unlabeled data using an incomplete taxonomy. The labeled data is directly used in an existing loss function (e.g., cross entropy). For the unlabeled data, we employ the available deepest concepts/nodes from the root, and the upper node is considered in case of a missing lower node. In this definition of the semantic loss, since there are no conflicts between different levels of concepts in the hierarchical taxonomy, there is no need for a sentence $\beta$ over $\textbf{B}$. The intuition behind this is to emphasize the information carried by unlabeled data and provide a level-based weighting for the incomplete taxonomy. 

\begin{small}
	\begin{equation}
	\label{eq:semi}
	L^s(\alpha,\beta,\textsf{p},\textsf{s}) \propto - \log \sum_{j \in \{1,k\}} \sum_{\textbf{x} \models \alpha} \prod_{i:\textbf{x} \models X_{i}} \textsf{p}_{i} \prod_{i:\textbf{x} \models \neg X_{i}} (1- \textsf{p}_{i})
	\end{equation}
\end{small}%

\begin{figure}[t]
	\centering
	\def\svgwidth{310pt}
	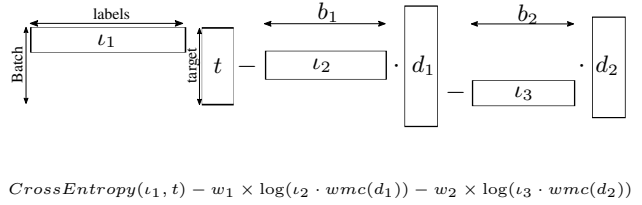
	\caption{An illustration for semi-supervised semantic loss. $\iota_1$ is the matrix of model output for the labeled (i.e., leaves in taxonomy) data, $\iota_2$ is the matrix of one-hot vectors over nodes in level $l_2$ without labels in leaves, and $\iota_3$ is the matrix of one-hot vectors over nodes in level $l_2$ without labels in leaves and level $l_2$.}
	\label{fig:illus_semi}
\end{figure}

Equation \ref{eq:semi} is illustrated in Figure \ref{fig:illus_semi}, including the training batch and SDDs.

In essence, Equation 1 and 2 expand semantic loss \cite{xu2018semantic} over hierarchical structure. They are proportional to the negative logarithm of the probability of generating a state that satisfies the constraint when sampling values according to $\textsf{p}$.

Our goal is to develop a tractable loss for computing both semantic loss and its gradient. From propositional logic theories, we know that a Model is a solution to a given propositional formula $\Delta$, and Model Counting or \#SAT is the problem of computing the number of models for $\Delta$. In case of mapping literals of the variables to non-negative real-valued weights, we will have Weighted Model Counting (WMC) \cite{chavira2008probabilistic,sang2005performing}. The well-known task of model counting  corresponds to the special case where all literal weights are 1 (and counts thus restricted to the natural numbers), whereas probabilistic inference (Prob) in a setting where all variables are independently assigned truth values at random restricts the weight function $\omega$ of WMC to values from [0, 1] such that weights of positive and negative literals for each var sum to one, i.e., for every variable $\upsilon$, $\omega(\upsilon) \in [0,1]$ and $\omega(\neg \upsilon)=1-\omega(\upsilon)$ \cite{kimmig2017algebraic}.

From \cite{darwiche2003differential}, we know about differential circuit languages that compute WMCs, which are amenable to backpropagation. Following \cite{xu2018semantic}, we use the circuit compilation techniques from \cite{darwiche2011sdd}, namely the Sentential Decision Diagram (SDD), to construct a Boolean circuit representing semantic loss. The SDD circuit form exhibits two main properties: determinism and decomposability, allowing us to compute both the values and gradients of the semantic loss in time linear to the size of the circuit \cite{darwiche2002knowledge}.

\subsection{GCN-based Approach}
\label{sec:gcn}

\begin{figure}
	\begin{subfigure}[b]{0.3\linewidth}
	\centering
	\scalebox{.75}{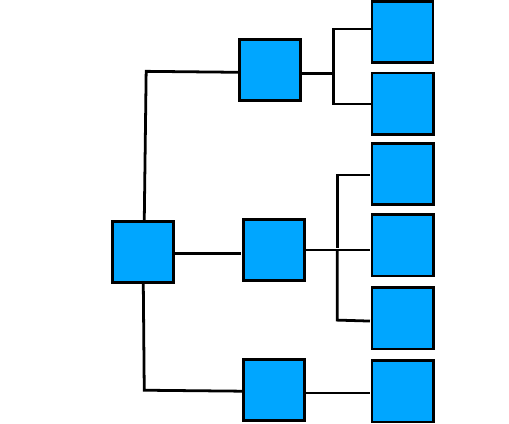}
	\centering\caption{A Hierarchical Taxonomy}
	\end{subfigure}
\hfill
	\begin{subfigure}[b]{0.5\linewidth}
	\centering
	\scalebox{.75}{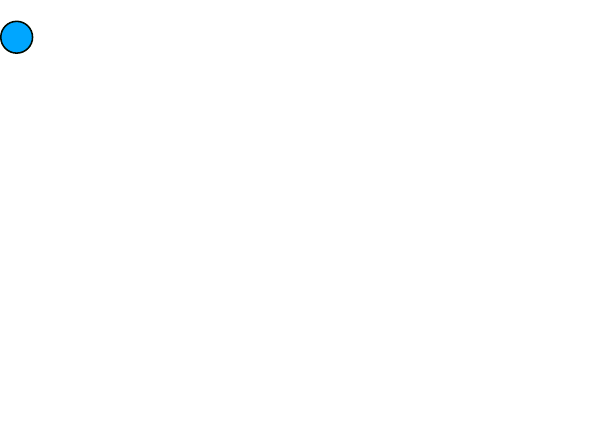}
	\centering\caption{A corresponding graph for the taxonomy in (a)}
	\end{subfigure}
\hfill
	\begin{subfigure}[b]{1.0\linewidth}
		\centering
		\scalebox{.8}{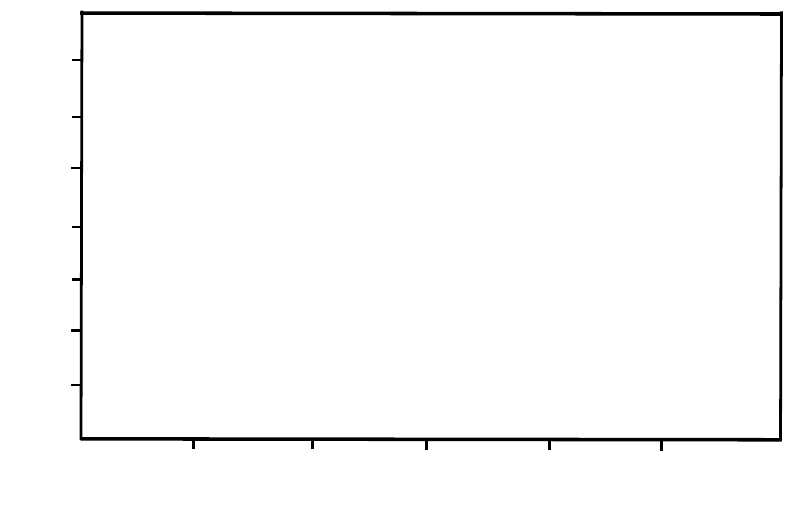}
		\caption{2-Dim plot of GCN representations for the nodes}
	\end{subfigure}

	\caption{An illustration for GCN encoding of the nodes in a taxonomy with labeled nodes $a_{0},..,a_{3}$, and unlabeled nodes $X_{1},..,X_{6}$.}
	\label{fig:gcn_rep}
\end{figure}

Graph Convolutional Networks is a powerful method presented for semi-supervised learning on graph-structured data \cite{kipf2016semi}, in which the authors introduced GCN to address the problem of classifying nodes, such as documents, in a graph, such as a citation network, where labels are only available for a small subset of nodes. Similarly, in representing hierarchical taxonomy in semi-supervised learning, we deal with the labeling concept in different levels of the hierarchy. Our objective is to identify representations for some nodes in the taxonomy, given the labels of other nodes. Moreover, the ability of GCN to handle symbolic inputs/outputs offers a differentiable alternative for semantic loss and logical constraints. These two reasons led us to utilize a graph neural network (GCN) for knowledge integration. We consider a hierarchical taxonomy as a labeled graph and seek the GCN encoding of any externally connected node to this graph. Figure \ref{fig:gcn_rep} provides a simple illustration of the nodes of a taxonomy with a 2-Dimensional encoding.

One issue with GCN is the large memory requirement when encoding a big graph-structured data to provide representations for each node. Moreover, using GCN on the entire graph data avoids the need for explicit regularization with another supervised loss function, such as Cross Entropy. In this section, we propose a method to incorporate the hierarchical taxonomy of a classification task as prior knowledge into the loss function through a Batch-based Graph Convolutional Networks (BGCN). A representation for a graph $A$ in GCN is defined as follows:

\begin{equation}
\label{eq:gcn}
H^{(l+1)}=\sigma (\tilde{D}^{- \frac{1}{2}} \tilde{A} \tilde{D}^{- \frac{1}{2}} H^{(l)} W^{(l)})
\end{equation}

\noindent where $\tilde{A}=A+L_{N}$ is the adjacency matrix of the undirected graph $A$ with added self-connections. $I_N$ is the identity matrix, $\tilde{D}_{ii}= \sum_{j} \tilde{A}_{ij}$  and $W^{(l)}$ is a layer-specific trainable weight matrix. $\sigma (.)$ Denotes an activation function (we used $ReLU$ in our experiments). $H^{(l)} \in \mathbb{R}$ is the matrix of activations in the $l^{th}$ layer; $H^{0}=X$, $H^{2}=softmax(H^{1})$.

\begin{figure}[t]
	\centering
	\includegraphics[scale=0.3]{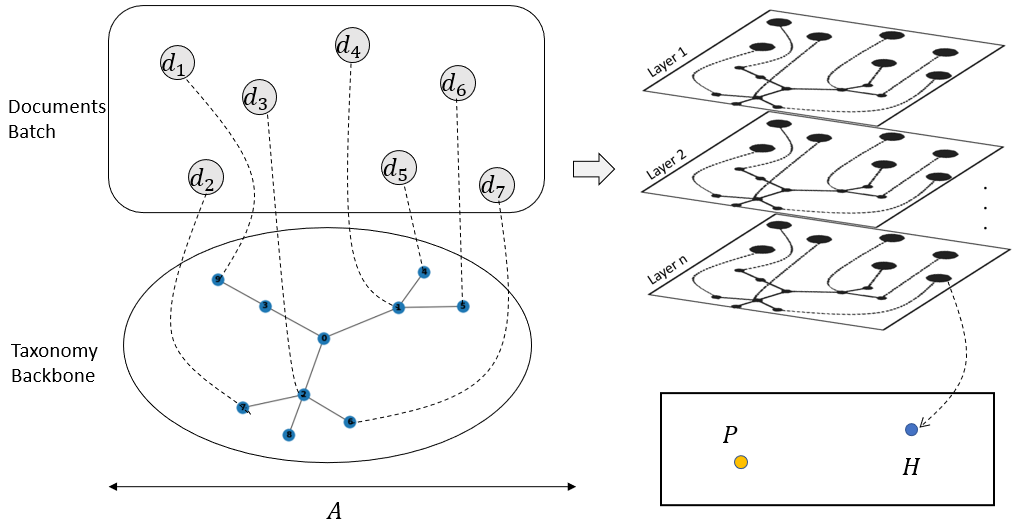}
	\centering\caption{An illustration of BGCN training workflow on documents $\mathbb{D}=\{d_1, d_2,..,d_7\}$ which are inter connected through a backbone graph, i.e., taxonomy.}
	\label{fig:gcn}
\end{figure}

We provide a taxonomy backbone graph for each batch, which is consistent across all batches and is generated from the taxonomy tree. Our method aims to mimic the Knowledge Distillation \cite{hinton2015distilling} approach, where the hierarchical taxonomy serves as knowledge transferred from a teacher model (i.e., GCN model) to a student model (i.e., Symbolic-based model). 
In this method, we want the training algorithm (DNN) to not only rely on a supervised loss function but also consider prior domain knowledge encoded in a taxonomy tree. Therefore, a batch includes a few documents from the training data along with the taxonomy tree representing the hierarchical categories of the classification, which serves as the backbone of a larger graph $A$. In other words, $A$ is a graph generated by connecting the documents of a batch to the taxonomy tree. The workflow of the end-to-end training of BGCN is shown in Figure \ref{fig:gcn}. The regularization term $\mathcal{L}_{reg}$, explicitly added to the existing loss function, is defined as:

\begin{equation}
\label{eq:lreg}
\mathcal{L}_{reg} = \parallel P - H \parallel^{2}_{2}
\end{equation}

The generated graph $A$ is used to provide representations $H$ for batch of document in the same space of the predicted probabilities from a supervised DNN. Euclidean distance is measured as the regularization loss to be added to the training loss. The final loss function is:

\begin{equation}
	\mathcal{L} = \mathcal{L}_{0} + w \times \mathcal{L}_{reg}
\end{equation}

\noindent where $\mathcal{L}_{0}$ is a Cross Entropy loss and $\mathcal{L}_{reg}$ is the regularization loss.

\section{Experimental Results}
\subsection{Taxonomy}
To demonstrate the impact of taxonomy on both fine-grained and general classification, we adopt a policy. Our aim is to incorporate a wide range of categories at the top-1 level (excluding the root), while avoiding excessive depth in the taxonomy hierarchy. Specifically, we limit the taxonomy to three levels, thereby preventing the task from becoming solely focused on fine-grained classification.

\subsection{Data}
\subsubsection{In-house Data:} We utilize two imbalanced Chinese datasets from a private company, each containing a large number of categories. The first dataset comprises user query logs from a Shopping Mall, consisting of 84 classes. The taxonomy associated with this dataset has three levels: level 1 encompasses 18 domains, level 2 comprises 45 intents, and level 3 includes 84 sub-intents. We split the data into training and testing datasets, consisting of 13,962 and 1,530 instances, respectively. The second dataset consists of user query logs from a Call Center Service, which consists of 134 classes. The hierarchical taxonomy for this dataset also has three levels: level 1 contains 5 domains, level 2 includes 24 intents, and level 3 (leaves/labels) contains 134 sub-intents. We split this dataset into training and testing data, comprising 19,214 and 1,619 instances, respectively.

\begin{figure}[t]
	\centering
	\begin{tikzpicture}
	\begin{axis}[
	ymin=0,
	ybar,
	bar width = 1pt,
	enlargelimits=0.01,
	width=.52\textwidth,
	height=0.25\textwidth,
	legend style={at={(0.4,0.4)}, anchor=north,legend columns=-1, font=\tiny}, 
	]
	\addplot[draw = red,
		line width = .2mm,
		fill = red,
		opacity=.6
	] table {geodist.dat};
	\legend{Call Center}
	\end{axis}
	\end{tikzpicture}
	\begin{tikzpicture}
	\begin{axis}[
	ybar,
	bar width = 1pt,
	enlargelimits=0.01,
	width=.50\textwidth,
	height=0.25\textwidth,
	legend style={at={(0.4,0.4)}, anchor=north,legend columns=-1, font=\tiny}, 
	]
	\addplot[draw = blue,
	line width = .2mm,
	fill = blue,
	opacity=.6
	] table {malldist.dat};	  
	\legend{Shopping Mall}
	\end{axis}
	\end{tikzpicture}
	\caption{Distribution of documents in different classes for In-house datasets.}
	\label{fig:bar}
\end{figure}
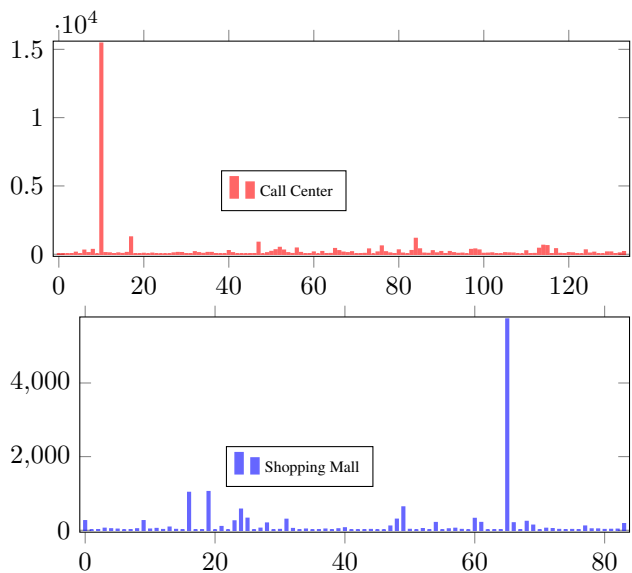

The documents in these datasets are short as they represent log queries. The average query length for the Shopping Mall dataset is 12.14, while for the Call Center Service dataset, it is 15.64. The distribution of documents across the classes is illustrated in Figure \ref{fig:bar}. 

\subsubsection{RCV1:} The Reuters Corpus Volume I (RCV1) is a widely recognized archive of over 800,000 manually categorized newswire stories provided by Reuters, Ltd. In our experiment, we focus on the Topic category, which has been extensively used in research with the Reuters corpus \footnote{\url{https://link.springer.com/content/pdf/bbm:978-3-642-04533-2/1.pdf}}. The Topic category aligns well with our requirements for multi-class classification and features a well-designed hierarchical taxonomy. The stories in the Topic category are organized into four hierarchical groups: CCAT (Corporate/Industrial), ECAT (Economics), GCAT (Government/Social), and MCAT (Market).

A new version of RCV1(RCV1-v2/LYRL2004) was provided by \cite{lewis2004rcv1}, where they described the coding policy, quality control procedures, and necessary corrections made to remove erroneous data.
The authors also made available detailed per-category experimental results, along with corrected versions of category assignments and taxonomy structures, through online appendices.\footnote{\url{http://www.jmlr.org/papers/volume5/lewis04a/lyrl2004_rcv1v2_README.htm}}.

Regarding this version of RCV1, we have generated a new dataset and a corresponding hierarchical taxonomy suitable for multi-class classification. To accomplish this, we have removed all documents that belong to more than one category. Based on the updated dataset, which includes 523,050 documents, we have created a hierarchical taxonomy with three levels tailored for multi-class classification. The first level comprises four categories, the second level consists of 33 categories, and the third level (leaves/labels) includes 53 sub-categories.

\subsubsection{Amazon Product Review:} This dataset consists of reviews from Amazon spanning an 18-year period, with approximately 35 million reviews up until March 2013 \cite{mcauley2013hidden}. Since our focus is on multi-class flat classification, similar to the Reuters dataset, we filter out documents with multiple labels and create an adjusted hierarchical taxonomy accordingly. The updated taxonomy is organized into three levels: the first level includes 22 categories, the second level consists of 116 categories, and the third level (leaves/labels) encompasses 300 sub-categories.

\subsection{Baseline Method}
\subsubsection{No Constraint:} 
LLM (Language Model with Latent Variables) has demonstrated significant improvements in machine learning performance, particularly in classification problems. It has yielded remarkable results even in few/zero-shot learning scenarios. To compare our results, we consider LLM as the baseline without utilizing any constraints or taxonomy. It is a fine-tuned BERT model with two stacked layers on the \textit{pooler\_output}, employing \textit{tanh} and \textit{softmax} activation functions, respectively.

\subsubsection{One-hot Constraint:} 
Due to the implicit knowledge already embedded in the BERT model through pre-training on a large corpus, it possesses powerful fine-tuning capabilities even with limited samples. It is challenging to demonstrate the impact of explicit injected knowledge in a BERT-based model. Therefore, to emphasize the effectiveness of integrating taxonomy knowledge into machine learning, we also employ the exactly-one or one-hot constraint, as presented in \cite{xu2018semantic}. This constraint captures the encoding used in multi-class classification, stating that for a set of indicators $X = \{X_1,...,X_n\}$, one and exactly one of those indicators must be true, with the rest being false. The logical function for three variables can be expressed as $(x_1 \wedge \neg x_2 \wedge \neg x_3)\vee(\neg x_1 \wedge x_2 \wedge \neg x_3)\vee(\neg x_1 \wedge \neg x_2 \wedge x_3)$.
%In our experiments, we only consider One-hot constraint for supervised learning, as it has totally different policy for unlabeled data compare to our approach, i.e., explained in Section \ref{sec:evalmeasure}.

\subsection{Ablation Study}
We conduct an ablation study on our proposed approach. We refer to our base model as the Tax-based model and consider the following variant:

\subsubsection{Tax-L1-based:} This variant of our base model includes only the first level $l_1$ of the taxonomy. Essentially, this variant is equivalent to the One-hot Constraint, as it represents the same formula proposed in \cite{xu2018semantic}. In this model variant, all upper levels of the taxonomy are removed, and only the leaf nodes are considered. To ensure a fair comparison with the One-hot Constraint, we evaluate the results only in a supervised fashion. To ensure reproducibility of our evaluation, we set all the \textit{seeds} and run for one epoch. The average performance improvements, measured by three metrics across all four datasets, are presented in Table \ref{tab:sup-one}.
%\subsubsection{Tax-L2-based:} is our base model that only includes second level of the taxonomy.

\subsection{Evaluation Measure}
\label{sec:evalmeasure}
We use three measures: \textit{Accuracy}, \textit{Macro Average F1-score}, and \textit{Weighted Average F1-score} to evaluate the obtained results. Specifically, for evaluating the effect of hierarchical taxonomy in imbalanced classification, we use \textit{Macro Average F1-score}, which is the arithmetic mean of F1-scores per class. It does not use weights (i.e., number of true labels of each class) for aggregation of F1-scores per class, and this results in a bigger penalization when a model does not perform well with the minority classes.%, i.e., exactly what happens in imbalance classification.

In all experiments, we obtain the results with/without injecting taxonomy into the existing data-driven loss function. We run all experiments for 10 epochs, with a batch size of 32. The experiments are repeated 3 times, and the best result for each method and baseline is selected.

\begin{table}
	\centering
	\resizebox{.49\textwidth}{!}{
	\begin{tabular}{lccc}
		\toprule
		Method               & Accuracy\% & Macro Avg F1\% & Weighted Avg F1\% \\
		\midrule
		Baseline       		 & 74.54           & 56.55         & 75.63  \\
		%Semantic Loss  		 & 74.98           & 58.49         & 75.18  \\
		+Symbolic-based       & \textbf{76.23}        & \textbf{60.84}          & \textbf{77.39} \\
		+GCN-based            & 76.10        & 59.46         & 77.32  \\
		\bottomrule
	\end{tabular}}
	\caption{A comparison of the methods in Supervised fashion on Call Center Service dataset.}
	\label{tab:sup-geo}
\end{table}

\begin{table}
	\centering
	\resizebox{.49\textwidth}{!}{
		\begin{tabular}{lccc}
			\toprule
			Method               & Accuracy\% & Macro Avg F1\% & Weighted Avg F1\% \\
			\midrule
			Baseline       		 & 93.80           & 81.69         & 93.57  \\
			%Semantic Loss		 & 93.93           & 82.13         & 93.68  \\
			+Symbolic-based      & \textbf{94.12}        & 83.60          & \textbf{93.79} \\
			+GCN-based           & 93.99        & \textbf{86.11}         & 93.74  \\
			\bottomrule
	\end{tabular}}
	\caption{A comparison of the methods in Supervised fashion on Shopping Mall dataset.}
	\label{tab:sup-mall}
\end{table}

\begin{table}
	\centering
	\resizebox{.49\textwidth}{!}{
		\begin{tabular}{lccc}
			\toprule
			Method               & Accuracy\% & Macro Avg F1\% & Weighted Avg F1\% \\
			\midrule
			Baseline       		 & 94.10           & 81.22         & 93.99  \\
			+Symbolic-based      & \textbf{95.16}           & 82.45         & \textbf{95.08} \\
			+GCN-based           & 94.49           & \textbf{83.55}         & 94.41  \\
			\bottomrule
	\end{tabular}}
	\caption{A comparison of the methods in Supervised fashion on Reuters dataset.}
	\label{tab:sup-reuters}
\end{table}

\begin{table}
	\centering
	\resizebox{.49\textwidth}{!}{
		\begin{tabular}{lccc}
			\toprule
			Method               & Accuracy\% & Macro Avg F1\% & Weighted Avg F1\% \\
			\midrule
			Baseline       		 & 52.62           & 41.67        & 51.78  \\
			+Symbolic-based      & \textbf{53.77}           & \textbf{42.14}        & \textbf{52.64} \\
			+GCN-based           & 53.58           & 41.71        & 52.58  \\
			\bottomrule
	\end{tabular}}
	\caption{A comparison of the methods in Supervised fashion on Amazon dataset.}
	\label{tab:sup-amazon}
\end{table}

\begin{table}
	\centering
	\resizebox{.49\textwidth}{!}{
		\begin{tabular}{lccc}
			\toprule
			Method               & Accuracy\% & Macro Avg F1\% & Weighted Avg F1\% \\
			\midrule
			Tax-based$_{Symbolic-based}$       		 & \textbf{+2.12}           & \textbf{+3.66}         & \textbf{+2.11}  \\
			Tax-based$_{GCN-based}$       		     & +0.59           & +2.72         & +0.71  \\
			Tax-L1-based            				 & +0.07           & +1.09         & +0.06  \\
			\bottomrule
	\end{tabular}}
	\caption{The results of the ablation study.}
	\label{tab:sup-one}
\end{table}

\begin{table}
	\centering
	\resizebox{.49\textwidth}{!}{
		\begin{tabular}{llccc}
			\toprule
			Portion	&		Method               & Accuracy\% & Macro Avg F1\% & Weighted Avg F1\% \\
			\midrule
			20\%	&		Baseline       		 & 65.94           & 43.58         & 64.95  \\
					&		+Symbolic-based      & \textbf{70.51}        & \textbf{52.85}          & \textbf{70.12} \\
					&		+GCN-based           & 68.00        & 50.28         & 67.74  \\
			\midrule
			30\%	&		Baseline       		 & 66.07    & 46.93      & 65.13  \\
					&		+Symbolic-based      & 69.09        & 54.29         & 69.04 \\
					&		+GCN-based           & \textbf{70.38}        & \textbf{54.68}         & \textbf{70.25}  \\
			\midrule
			40\%	&		Baseline       		 & 68.58           & 51.67         & 68.20  \\
					&		+Symbolic-based      & \textbf{71.93}        & \textbf{58.54}         & \textbf{71.88} \\
					&		+GCN-based           & 69.99        & 55.17         & 69.59  \\
			\bottomrule
	\end{tabular}}
	\caption{A comparison of the methods in Semi-Supervised fashion on Call Center Service dataset.}
	\label{tab:semi-geo}
\end{table}

\begin{table}
	\centering
	\resizebox{.49\textwidth}{!}{
		\begin{tabular}{llccc}
			\toprule
		Portion	&		Method               & Accuracy\% & Macro Avg F1\% & Weighted Avg F1\% \\
			\midrule
		20\%	&		Baseline       		 & 87.67           & 57.54         & 86.93  \\
				&		+Symbolic-based      & \textbf{90.88}        & 66.12         & \textbf{90.09} \\
				&		+GCN-based           & 90.56        & \textbf{67.18}         & 89.84  \\
			\midrule
		30\%	&		Baseline       		 & 91.53           & 71.87         & 90.48  \\
				&		+Symbolic-based      & \textbf{92.22}        & 76.23         & 91.75 \\
				&		+GCN-based           & \textbf{92.22}        & \textbf{81.26}         & \textbf{91.93}  \\
			\midrule
		40\%	&		Baseline       		 & 92.22           & 75.70         & 91.73  \\
				&		+Symbolic-based      & \textbf{92.68}        & 76.48         & \textbf{92.20} \\
				&		+GCN-based           & 92.22        & \textbf{76.78}         & 91.83  \\
			\bottomrule
	\end{tabular}}
	\caption{A comparison of the methods in Semi-Supervised fashion on Shopping Mall dataset.}
	\label{tab:semi-mall}
\end{table}

\begin{table}
	\centering
	\resizebox{.49\textwidth}{!}{
		\begin{tabular}{llccc}
			\toprule
			Portion	&		Method               & Accuracy\% & Macro Avg F1\% & Weighted Avg F1\% \\
			\midrule
			20\%	&		Baseline       		 & 91.23           & 68.51         & 90.94  \\
					&		+Symbolic-based      & 91.60	       & 71.88         & 91.38 \\
					&		+GCN-based           & \textbf{91.70}           & \textbf{74.94}         & \textbf{91.43}  \\
			\midrule
			30\%	&		Baseline       		 & 92.74           & 75.38         & 92.56  \\
					&		+Symbolic-based      & \textbf{93.43}           & 76.54         & \textbf{93.40} \\
					&		+GCN-based           & 92.54           & \textbf{77.08}         & 92.45  \\
			\midrule
			40\%	&		Baseline       		 & 93.26           & 76.52         & 93.09  \\
					&		+Symbolic-based      & \textbf{93.53}           & 79.93         & \textbf{93.42} \\
					&		+GCN-based           & 93.48           & \textbf{80.26}         & 93.35  \\
			\bottomrule
	\end{tabular}}
	\caption{A comparison of the methods in Semi-Supervised fashion on Reuters dataset.}
	\label{tab:semi-reuters}
\end{table}

\begin{table}
	\centering
	\resizebox{.49\textwidth}{!}{
		\begin{tabular}{llccc}
			\toprule
			Portion	&		Method               & Accuracy\% & Macro Avg F1\% & Weighted Avg F1\% \\
			\midrule
			20\%	&		Baseline       		 & 40.93           & 19.99         & 36.94  \\
					&		+Symbolic-based      & \textbf{44.00}	       & \textbf{27.94}         & \textbf{41.83} \\
					&		+GCN-based           & 43.76           & 25.34         & 40.90  \\
			\midrule
			30\%	&		Baseline       		 & 44.66           & 26.37         & 42.34  \\
					&		+Symbolic-based      & 46.45           & \textbf{32.02}         & \textbf{44.89} \\
					&		+GCN-based           & \textbf{46.64}           & 29.97         & 44.40  \\
			\midrule
			40\%	&		Baseline       		 & 48.09           & 32.13         & 45.43  \\
					&		+Symbolic-based      & \textbf{49.14}           & \textbf{36.47}         & \textbf{47.87} \\
					&		+GCN-based           & 48.80           & 33.39         & 47.01  \\
			\bottomrule
	\end{tabular}}
	\caption{A comparison of the methods in Semi-Supervised fashion on Amazon dataset.}
	\label{tab:semi-amazon}
\end{table}

\begin{table}
	\centering
	\resizebox{.49\textwidth}{!}{
		\begin{tabular}{llccc}
			\toprule
			Dataset			&		\%      & Level-1 Masking rate & Level-2 Masking rate & Level-3 Masking rate \\
			\midrule
			Shopping Mall	&		20      & 0       & 0.6008       & 0.8025   \\
			&		30      & 0       & 0.4064       & 0.7023   \\
			&		40      & 0       & 0.2038       & 0.6070   \\
			\midrule                                                            
			Call Center		&		20      & 0       & 0.6	         & 0.8  	\\
			&		30      & 0       & 0.4        	 & 0.7		\\
			&		40      & 0       & 0.2	         & 0.6		\\
			\bottomrule
	\end{tabular}}
	\caption{The portion of masking in different levels of the taxonomies for Shopping Mall dataset.}
	\label{tab:semi-ratio}
\end{table}

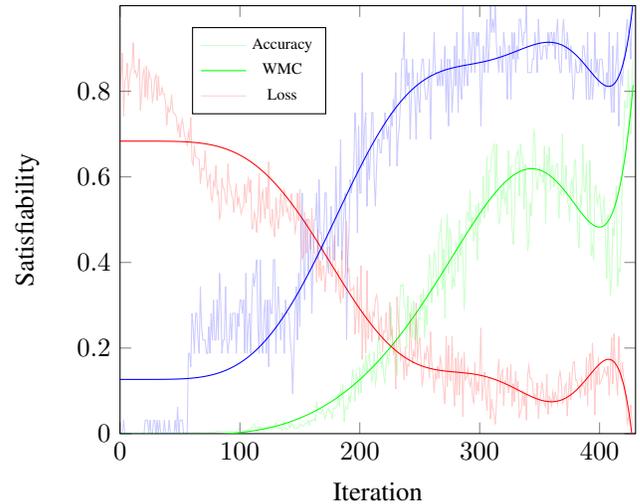
\begin{figure}[t]
	\begin{tikzpicture}
	\begin{axis}[
	xlabel={Iteration},
	ylabel={Satisfiability},
	xmin=0, xmax=430,
	ymin=0.0, ymax=1.0,
	xtick={0,100,200,300,400},
	ytick={.0, .2 , .4, .6, .8},
	legend style={at={(0.4,0.95)}, font=\tiny},
	]
	
	\addplot[
	color=green,
	opacity=.2
	]
	table {wmc.dat};
	\addplot[
	color=green,
	]
	table {wmcreg.dat};
	\addplot[
	color=red,
	opacity=.2
	]
	table {loss.dat};
	\addplot[
	color=red,
	]
	table {lossreg.dat};
	\addplot[
	color=blue,
	opacity=.2
	]
	table {acc.dat};
	\addplot[
	color=blue,
	]
	table {accreg.dat};
	\legend{Accuracy,WMC,Loss}
	\end{axis}
	
	\end{tikzpicture}
	\caption{The satisfiability of WMC in semantic loss, training loss, and accuracy in first epoch.}
	\label{fig:wmc}
\end{figure}

In semi-supervised learning, we define a policy to generate datasets in which some of the target labels, i.e., leaves in taxonomy, for documents and some of internal nodes in the taxonomies, are randomly masked/removed. Table \ref{tab:semi-ratio} shows the rates of masking in different levels of the taxonomies for Shopping Mall dataset used in our experiments. For example, 20\% semi-supervised learning means that in the leaves of the taxonomy, which also correspond to the target labels, 80\% of the data rely on the taxonomy. Moreover, from the 80\% unlabeled data in this case, 40\% of it relies on level 2 of the taxonomy, and the rest relies on first level. To provide a fair comparison, semi-supervised datasets with same portion of data (e.g., 20\%) in all experiments with different methods are identical.

To examine the pure effect of taxonomy, we do not consider the taxonomy knowledge for both the labeled and unlabeled data in semi-supervised learning (Tables \ref{tab:semi-geo}, \ref{tab:semi-mall}, \ref{tab:semi-reuters}, and \ref{tab:semi-amazon}). We only take the knowledge from taxonomy into account for unlabeled data, and separately in different experiments (Tables \ref{tab:sup-geo}, \ref{tab:sup-mall}, \ref{tab:sup-reuters}, and \ref{tab:sup-amazon}), it is considered for all data. 

Injecting the abstracted taxonomy into the existing loss function provides a learning signal on unlabeled samples by punishing the underlying learner to make decisions that satisfy the constraint.
Figure \ref{fig:wmc} shows the satisfiability of the regularization term (i.e., WMC in semantic loss) of the training loss function in the first epoch, together with the training loss and the accuracy. The results indicate that as we reason on the hierarchical taxonomy, we will see improvement in categorizing documents.

The significant effect of the taxonomy can be seen in the semi-supervised tables (Tables \ref{tab:semi-geo}, \ref{tab:semi-mall}, \ref{tab:semi-reuters}, and \ref{tab:semi-amazon}), and clearly, it indicates that the main effort goes for the Macro average F1-score. 
Although \textit{Symbolic-based} method shows better results, specially in accuracy of learner in supervised and semi supervised manners. But, as it is shown in Tables \ref{tab:sup-mall}, \ref{tab:sup-reuters}, \ref{tab:semi-mall}, and \ref{tab:semi-reuters} \textit{GCN-based} method shows better results in Macro Average F1-score. Size of the hierarchical taxonomies generated for Shopping Mall and Reuters datasets are smaller that other two datasets in our experiments, regarding a fixed size of batch, it indicates the effect of GCN in a long-tailed problem.
However, as shown in our experiments, the effect of the taxonomy on the model`s performance reduces increasing of the labeled samples.

As Table \ref{tab:sup-one} shows, the one-hot constraint does not significantly affect performance, and it is almost zero 
because the constraint is always satisfied by existing data-driven loss, and it can perfectly fit training data. 
Despite this satisfaction, we can still see the effect of the taxonomy on overall performance, specially in minor classes, even in fully supervised learning Table \ref{tab:sup-geo} and \ref{tab:sup-mall}. This is because of the hierarchical structure of the constraint, which alleviates model output distributions over the classes to allow conditioning on upper concepts.

\section{Conclusion}
In this paper, we aimed to explore several key challenges in deep learning: reasoning, semi-supervised learning, and the long-tailed issue. We developed two methods to represent and integrate a hierarchical taxonomy of labels into the loss function of a flat classifier. We demonstrated the effect of these methods in supervised and semi-supervised learning. Moreover, our experimental evaluations show that integrating a well-designed hierarchical taxonomy into the learning algorithm of a neural network effectively guides the learner to achieve significant results on long-tailed problems.
%If you are dealing with a general classification task, and your dataset is imbalance, then if you are able to generate a 3-level taxonomy for your labels, then you will get improvement in your accuracy, mainly in minority classes.

\appendix

%% The file kr.bst is a bibliography style file for BibTeX 0.99c
\bibliographystyle{kr}
\bibliography{kr-sample}

\end{document}